\documentclass{article}
\usepackage{spconf,amsmath,graphicx}

\title{$F_0$ Modeling in HMM-Based Speech Synthesis System using Deep Belief Network}
\twoauthors
  {Sankar Mukherjee}
	{Yantra Software\\
	Hyderabad\\
	India\\
sankar1535@gmail.com}
  {Shyamal Kumar Das Mandal}
	{Centre for Educational Technology\\
Indian Institute of Technology Kharagpur\\
Kharagpur, India\\
sdasmandal@cet.iitkgp.ernet.in}
\begin{document}
%
\maketitle
\begin{abstract}
In recent years multilayer perceptrons (MLPs) with many hidden layers – Deep Neural Network (DNN) has performed surprisingly well in many speech tasks, i.e. speech recognition, speaker verification, speech synthesis etc. Although in the context of $F_0$ modeling these techniques has not been exploited properly.  In this paper, Deep Belief Network (DBN), a class of DNN family has been employed and applied to model the $F_0$ contour of synthesized speech which was generated by HMM-based speech synthesis system. The experiment was done on Bengali language. Several DBN-DNN architectures ranging from four to seven hidden layers and up to 200 hidden units per hidden layer was presented and evaluated. The results were compared against clustering tree techniques popularly found in statistical parametric speech synthesis. We show that from textual inputs DBN-DNN learns a high level structure which in turn improves $F_0$ contour in terms of objective and subjective tests.
\end{abstract}
\section{Introduction}
\label{sec:intro}

Prosody plays the most important role in generating natural and intelligent speech. Prosody is a collection of supra-segmental features (duration, intonation, co-articulation pattern) which contributes additional information to speech that do not found in text. In prosody, $F_0$ contour has a significant contribution that is crucial to human speech perception. Knowledge of these parameters are implicit in speech signal i.e. it's hard to capture the rules governing these knowledge.  Traditional shallow architectures (i.e. statistical model with few levels of computation units) are fine in limited domain but they are not capable of handling these parameters when there are lots of variations in the test set. In this context, DNN is famous for its ability to capture internal representations that become increasingly complex. 

In recent years there has been a lot of interest in applying DNN in different speech processing tasks. Main advantage of these deep learning techniques that they can learn from unlabeled data and only a limited number of labeled example are needed to fine-tune the model according to the specific tasks at hand. DNNs are basically a multilayer perceptron with many hidden layers. DBN is a class of DNN family. It is a probabilistic generative model with multiple layers of stochastic, hidden variables. Each pair of layers is treated as a Restricted Boltzmann Machine (RBM) which is a bipartite undirected graphical model with two-layer architecture. The training of DBN as described in \cite{hinton2006fast} is to first initialize the weights of each layer greedily in a purely unsupervised way and then fine-tune all the weights jointly to further improve the likelihood. The resulting DBN is considered as a hierarchy of non-linear feature detectors that can learn complex statistical patterns. Rather than initialize random weights in a DNN, the weights learned by DBN can be used as the weights of a DNN. This is commonly called \textit{pre-training} of DNN. The whole DNN can be further fine-tune by a small number of labeled training data. DBN-DNN (DNN pre-trained by DBN and fine-tune by labeled data) is successfully applied in speech, audio, image and text data \cite{mnih2010learning} \cite{collobert2008unified}. This advances triggered interest in applying deep learning techniques in speech synthesis tasks. In recent years, DBNs have been successfully applied to modeling speech signals, such as spectrogram coding \cite{deng2010binary}, speech recognition \cite{dahl2012context}, and acoustic-articulatory inversion mapping \cite{uria2011deep}, where they mainly act as the pre-training methods for a deep autoencoder or a deep neural network (DNN). In statistical parametric speech synthesis domain DBNs have also been studied very recently \cite{zenstatistical} \cite{kang2013multi} \cite{ling2013modeling}.

Use of DBN to model $F_0$ contour \cite{kang2013multi} \cite{fernandez2013f0} was not new. But in \cite{fernandez2013f0} DBN was used as feature extractor for Gaussian Process Regression which is a non-parametric model. Our work was different in the sense that our prediction model was based on DNN with weight initialized by DBN. This makes the proposed model completely parametric which has many advantages like smaller foot print in contrast to non-parametric model. By doing that the proposed method could be used in hand held devices as a stand alone application. Another aspect was rather than discontinues $F_0$ contour \cite{kang2013multi} we train the DNN with continues $F_0$ contour which adds simplicity to the model.

This paper organized as follows. In Section 2, we will briefly review the basic techniques of RBMs and DBNs. In Section 3, we will describe the details of our proposed method. Section 4 reports our experimental results. Section 5 gives the conclusion and the discussion on our future work.

\section{Architecture of Deep Belief Network}
\label{sec:format}

\subsection{Restricted Boltzmann Machines}
RBM is a special type of Markov random field that has one layer of (Bernoulli) stochastic hidden units and one layer of (Bernoulli or Gaussian) stochastic visible or observable units. There are no visible-visible or hidden-hidden connections but all the visible units are connected to all the hidden units.The weights between the connections of the visible units v and hidden units h define a probability distribution over the visible units v  via an energy function \cite{welling2004exponential}. Depending on the visible unit (i.e. Bernoulli or Gaussian) there are two types of energy function (i.e. Bernoulli (visible)-Bernoulli (hidden) and Gaussian (visible)-Bernoulli (hidden)) of the joint configuration (v,h). Gaussian-Bernoulli RBMs is used to convert real-valued stochastic variables into to binary stochastic variables, which then further processed using the Bernoulli-Bernoulli RBMs. In this work Bernoulli-Bernoulli is used and its energy function is defined as 
\begin{equation}\label{bb} 
E(\textbf{v},\textbf{h};\theta) = - \sum_{i = 1}^{V}\sum_{j = 1}^{H} w_{ij}v_i h_j - \sum_{i = 1}^{V} b_i v_i - \sum_{j = 1}^{H} a_j h_j
\end{equation}
Where $\theta$ = (\textbf{w}, \textbf{b}, \textbf{a}) and $ w_{ij} $ represents the symmetric interaction term between visible unit $v_i$ and hidden unit $h_j$, $b_i$ and $a_j$ the bias terms, and V and H are the numbers of visible and hidden units. The joint distribution p(v,h;$\theta$) over the visible units v and hidden units h, given the model parameters $\theta$, in terms of an energy function E(v,h;$\theta$)  is defined as
\begin{equation}
p(\textbf{v},\textbf{h};\theta) = \frac{exp (-E(\textbf{v},\textbf{h};\theta))}{Z}
\end{equation}
where $Z = \Sigma_v \Sigma_h exp(-E(\textbf{v},\textbf{h};\theta))$ is a normalization factor. The marginal probability that the model assigns to a visible vector \textbf{v} is
\begin{equation}
p(\textbf{v};\theta) = \frac{\Sigma_h exp (-E(\textbf{v},\textbf{h};\theta))}{Z}
\end{equation}
As there are no hidden-hidden or visible-visible connections, the conditional distributions are factorial and are given by
\begin{equation}
p(h_j = 1 | \textbf{v};\theta) = \sigma (\sum_{i = 1}^{V} w_{ij} v_i + a_j) 
\end{equation}
\begin{equation}
p(v_j = 1 | \textbf{h};\theta) = \sigma (\sum_{i = 1}^{H} w_{ij} h_i + b_i)
\end{equation}
where $\sigma(x)$ is the activation function. Here $\sigma(x) = \frac{1}{(1+exp(x))}$ was considered. Taking the stochastic gradient descent of the negative log likelihood \textit{l} the update rule for the RBM weights as
\begin{equation}
\Delta w_{ij}(t+1) = m\Delta w_{ij}(t) - \alpha\frac{\partial l}{\partial w_{ij}}
\end{equation}
where $\alpha$ is the learning rate and \textit{m} is the momentum factor used to smooth out the weight updates. General form of the derivative
of the log likelihood of the data can be written as
\begin{equation}
- \frac{\partial l(\theta)}{\partial w_{ij}} = E_{data}(v_i,h_j) - E_{model}(v_i,h_j)
\end{equation}
where $E_{data}(v_i,h_j)$ is the expectation observed in the training set and $E_{model}(v_i,h_j)$ is that same expectation under the distribution defined by the model. But, $E_{model}(v_i,h_j)$ is computationally very expensive to compute so the contrastive divergence (CD) algorithm \cite{hinton2002training} to the gradient is used where $E_{data}(v_i,h_j)$ is replaced by running the Gibbs sampler initialized at the data for one full step.

%
%

\subsection{Deep Belief Network}
A DBN is formed by Stacking a number of the RBMs learned layer by layer from bottom up. 
In this model, each layer captures the correlations among the activities of hidden features in the layer below.  The top two layers of the DBN form an undirected bipartite graph. The lower layers form a directed graph with a top-down direction to generate the visible units. Given the training samples of the visible units, it is difficult to estimate the model parameters of a DBN directly under the maximum likelihood criterion due to the complex model structure with multiple hidden layers. Therefore, a greedy learning algorithm has been proposed and popularly applied to train the DBN in a layer-by-layer manner \cite{hinton2006fast}. After learning Bernoulli-Bernoulli RBM the activation probabilities of its hidden units was treated as the data for training the Bernoulli-Bernoulli RBM one layer up. The activation probabilities of the $2^{nd}$-layer Bernoulli-Bernoulli RBM are then used as the visible data input for the $3^{rd}$-layer Bernoulli-Bernoulli RBM, and so on. This greedy procedure above achieves approximate maximum likelihood learning. It has been proved that this greedy learning algorithm can improve the lower bound on the log-likelihood of the training samples by adding each new hidden layer \cite{hinton2006fast} \cite{salakhutdinov2009learning}. AIS-based partition function estimation with approximate inference \cite{salakhutdinov2009learning} used to estimate the lower bound on the log-probability. 

\section{Proposed $F_0$ Modeling Approach}
\begin{figure}[!ht]
\centering
\includegraphics[scale=0.2]{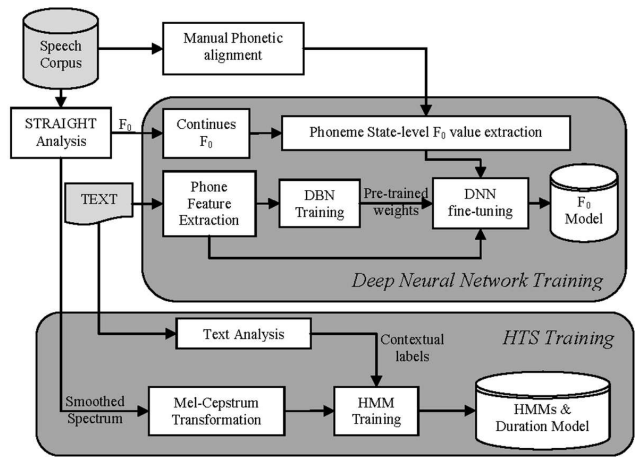}
\caption{Training stage of the proposed $F_0$ method}
\label{fig:traning}
\end{figure}

As shown in Figure \ref{fig:traning}, a database of speech and corresponding text sentences was used as the training corpus. CRBLP speech corpus \cite{alam2010development} is employed here for the whole experiment and it consists of one male voice of age 27. STRAIGHT \cite{kawahara1997speech}, a high-quality analysis and synthesis algorithm, was adopted to estimate the spectrum and $F_0$ contours with 10-ms frame rate. Continues $F_0$ contour was formed using step described in \cite{narusawa2002method} which resulted an approximation of original $F_0$ contour consisting of third order polynomial segments. For the input of the neural network a set of textual features were extracted from the raw text. Table \ref{textfeat} illustrates the features considered for this work. The phoneme consists of 30 consonants and 16 vowels. Max syllable length 6 and max 10 syllable words was considered and which was sufficient for Bengali language. These features were re-encoded using One-of-N codes which resulted 220 binary features. All though the phoneme and syllable properties may differ for language to language. To mitigate this language dependency problem we only need to adapt language specific text analysis module. Apart from the text analysis module the whole system is completely independent of Language. 

\begin{table}[h]
\centering
  \caption{\label{textfeat} {\it Textual features extracted from raw text}}
\vspace{2mm}
\begin{tabular}{|l|r|}
\hline
\multicolumn{1}{|c|}{\textbf{Feature Name}}                                                        & \multicolumn{1}{c|}{\textbf{No. of Features}} \\ \hline
\begin{tabular}[c]{@{}l@{}}Phoneme identity \\ {[}Previous/Current/Next{]}\end{tabular}            & 46 * 3 = {[}138{]}                            \\ \hline
No. of syllable in current word                                                                    & {[}10{]}                                      \\ \hline
\begin{tabular}[c]{@{}l@{}}Phoneme position in syllable \\ {[}Forward/Backward{]}\end{tabular}     & 6 * 2 = {[}12{]}                              \\ \hline
\begin{tabular}[c]{@{}l@{}}Syllable position in word \\ {[}Forward/Backward{]}\end{tabular}        & 10 * 2 = {[}20{]}                             \\ \hline
\begin{tabular}[c]{@{}l@{}}No. of phonemes in syllable \\ {[}Previous/Current/Next{]}\end{tabular} & 6 * 3 = {[}18{]}                              \\ \hline
\begin{tabular}[c]{@{}l@{}}Vowel position in syllable\\  {[}Forward{]}\end{tabular}                & {[}6{]}                                       \\ \hline
Vowel identity in the syllable                                                                     & {[}16{]}                                      \\ \hline
\textbf{Total}                                                                                     & \textbf{220}                                  \\ \hline
\end{tabular}
\end{table}

\subsection{DBN Training}
Input to the DBN was binary i.e. we used Bernoulli-Bernoulli RBM. From the corpus 7000 sentences are chosen to train the DBN. Each DBN layer was pre-trained for 50 epochs as a RBM with mini-batch of size 10. Average gradients were computed on the mini-batches and parameters were updated with a learning rate of 0.002 and a momentum of 0.95. Dif-ferent architecture of DBN illustrated in Figure 3 were trained with these configuration.

\subsection{DNN Training}
DNN is pre-trained by the DBN which means weights of the DNN is initialized from the trained DBN. In order to fine-tune the DNN 1000 sentences (excluding the previous 7000) are taken from the CRBLP corpus and they are phonetically aligned by HTK toolkit (5-state HMM). Finally the phonetic boundaries are manually corrected. From the 1000 sentences 500 are chosen to train, 200 to cross validation, 300 to test the DNN. Output of the DNN are log $F_0$ values corresponds to each phoneme state. This is because sentences are segmented using 5-state HMMs which results 5 states for each phoneme or each observation corresponds to roughly 1/5 of the phonemes. $F_0$ values corresponding to each state were calculated from the continues $F_0$ contour with the duration information generated by HTK. These $F_0$ values act as the output of DNN. We used mean squared error as the objective function with sparsity target of 0.05 and 0.002 weight decay. With mini-batches of 100 states backpropagation training was evaluated using a cross validation set. Loss is measured for the 10 epochs and if loss increased then learning rate was decreased by a factor of two.

The experiment was carried out on DELL precision T3600 workstation which is a 6 core computer with a CPU clock speed of 3.2 GHz, 12MB of L3 cache and 64GB DDR3 RAM. The training also used an NVIDIA Quadro 4000 general purpose graphical processing unit (GPGPU) for matrix multiplication.
\begin{figure}[!ht]
\centering
\includegraphics[scale=0.2]{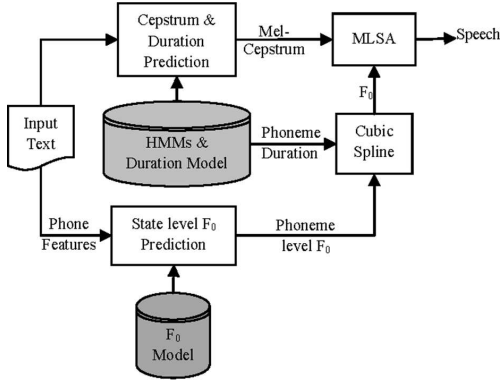}
\caption{Synthesis stage of the proposed method}
\label{fig:prediction}
\end{figure}

\subsection{Synthesis Stage}
Figure \ref{fig:prediction} shows the synthesis procedure of the proposed system. It has two main parts - generation of the cepstrum and duration, and prediction of the $F_0$ values. At the synthesis time the same features were extracted corresponding to each phoneme and they were fed to the DNN. With the previously trained weights DNN then predict the $F_0$ values. From the HTS phoneme state-duration model the duration of each states were extracted. Cubic spline interpolation were performed on the state $F_0$ values with the duration information so that the resulted $F_0$ has the same length. As a result, a continuous pitch contour was generated. Using MLSA filter synthesized speech was generated.

\section{Results and Evaluation}
Objective evaluation are conducted to evaluate the performance of the different DBN-DNN model. The best model is chosen to further be evaluated using subjective tests. In order to compare the proposed model against the clustering tree with multi-space probability distribution (MSD-HMM) $F_0$ model included in the HTS synthesis engine toolkit \cite{hts}, a Bengali-HTS \cite{mukherjee2012bengali} system is constructed. It is built with HTS 2.2 with $34^{th}$ order MGC coefficient, 10ms Blackman window, 5ms shift and 0.53 frequency wrapping factor.

\subsection{Objective Evaluation and Model Selection}
\begin{figure}[!ht]
\centering
\includegraphics[scale=0.3]{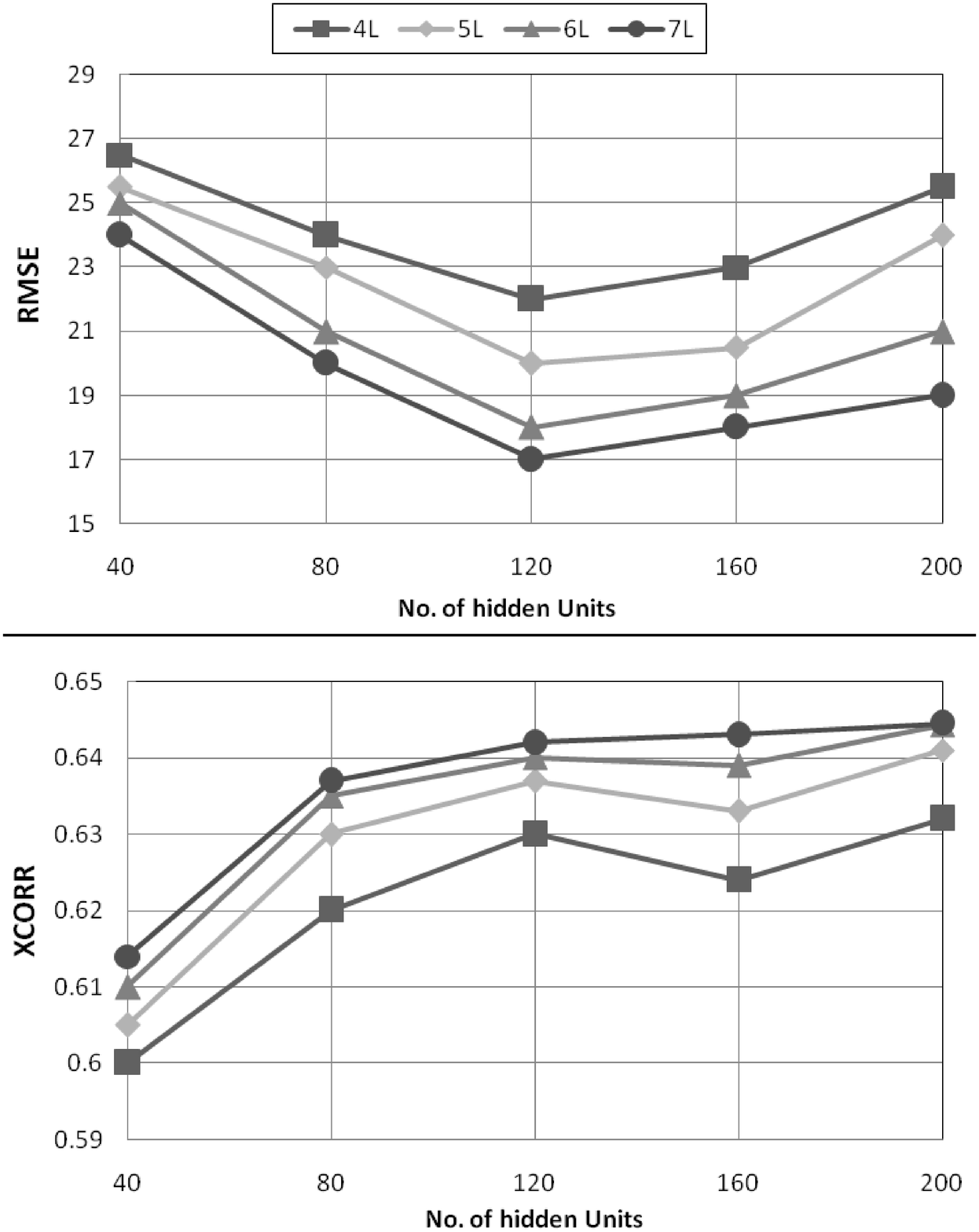}
\caption{RMSE and XCORR of the predicted $F_0$ on the test set}
\label{fig:rmse_xcorr}
\end{figure}

For the objective evaluation two types of metrics i.e. cross-correlation (XCORR) and root mean square error (RMSE) have been calculated. Several architecture of DNN has been constructed and their performance is evaluated using these two metrics. Figure \ref{fig:rmse_xcorr} illustrates the different RMSE and XCORR values of the predicted $F_0$ values on the test set (300 sentences). From the Figure \ref{fig:rmse_xcorr} it is clear that XCORR is improving on the test set with increasing hidden layer size but the performance of RMSE is decreasing. So we choose \textbf{DBN-DNN (120U-7L)} which yields the best performance according to the two metrics (RMSE 17 and XCORR 0.64). Table \ref{objective}  presents the objective test results performed on the test set between MSD-HMM and DBN-DNN. \textbf{DBN-DNN (120U-7L)} is selected for subjective evaluation.

\begin{table}[h]
\centering
\caption{\label{objective} {\it Result of objective test between MSD-HMM and DBN-DNN}}
\vspace{2mm}
\begin{tabular}{|c|c|c|}
\hline
               & \textbf{MSD-HMM} & \textbf{DBN-DNN} \\ \hline
\textbf{RMSE}  & 25.03            & 17               \\ \hline
\textbf{XCORR} & 0.49             & 0.64             \\ \hline
\end{tabular}
\end{table}

\subsection{Subjective Evaluation}
\begin{figure}[!ht]
\centering
\includegraphics[scale=0.3]{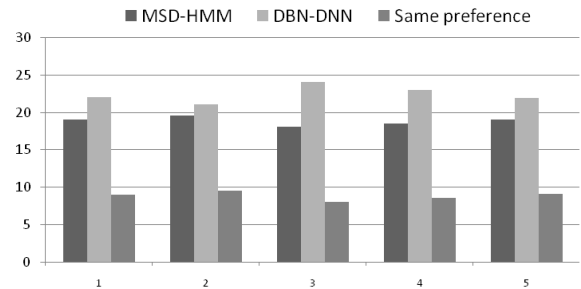}
\caption{ABX score of the of the two systems}
\label{fig:abx}
\end{figure}

For subjective measurement, ABX is performed with 5 subjects (3 male, 2 female). All subjects are not speech experts and native speakers of Bengali. In this experiment, the 50 test sentences are synthesized using DBN-DNN (120U-7L) and MSD-HMM. Participants are asked to choose their preferred one. It can be seen from Figure \ref{fig:abx} that 82\% of the time subjects has a preference towards one of the two systems and majority (54\% vs 46\%) preferred proposed system and it is statistically significant with p $<$ 0.001.

\section{Conclusion \& Future Works}
In this work we have applied Deep Belief Network (DBN) to model the $F_0$ contour of synthesized speech which is generated by HMM-based Speech Synthesis System. DBN acted as a high level feature extractor from the raw input text. Neural network is trained for each phoneme properties i.e. for the input to the neural network textual features (phoneme identity, syllable counts etc.) and for the output normalized log $F_0$ values are used. Although the whole experiment is conducted on Bengali Language but it can be applied to any languages. From the objective metrics and subjective test it is found that proposed model has more preference than MSD-HMM based model which is found in many standard text-to-speech synthesis systems.

\bibliographystyle{IEEEbib}
\bibliography{IEEEbib}
\end{document}